%% file: main.tex
\documentclass[10pt,twocolumn,letterpaper]{article}

\usepackage{multirow}
\usepackage{makecell}
\usepackage{subfigure}
\usepackage[table]{xcolor}
\definecolor{lightpink}{RGB}{255, 182, 193}
\usepackage[pagenumbers]{iccv} 


\definecolor{iccvblue}{rgb}{0.21,0.49,0.74}
\usepackage{subcaption}
\usepackage{float}
\usepackage[pagebackref,breaklinks,colorlinks,allcolors=iccvblue]{hyperref}

\def\paperID{*****} 
\def\confName{ICCV}
\def\confYear{2025}

\title{PP-FormulaNet: Bridging Accuracy and Efficiency in Advanced Formula Recognition}

\author{Hongen Liu\textsuperscript{1,2}, Cheng Cui\textsuperscript{1}, Yuning Du\textsuperscript{1}, Yi Liu\textsuperscript{1}, Gang Pan\textsuperscript{2} \\
\textsuperscript{1}PaddlePaddle Team, Baidu Inc.\\
\textsuperscript{2}College of Intelligence and Computing, Tianjin University\\
\tt\small \{liuhongen, cuicheng01\} @baidu.com, \tt\small pangang@tju.edu.cn
}

\begin{document}
\maketitle
\input{sec/0_abstract}    
\input{sec/1_intro}
{
    \small
    \bibliographystyle{unsrt}
    \bibliography{main}
}
\input{sec/X_suppl}

\end{document}

%% file: sec/0_abstract.tex
\begin{abstract}
Formula recognition is an important task in document intelligence. It involves converting mathematical expressions from document images into structured symbolic formats that computers can easily work with. LaTeX is the most common format used for this purpose. In this work, we present \textbf{PP-FormulaNet}, a state-of-the-art formula recognition model that excels in both accuracy and efficiency. To meet the diverse needs of applications, we have developed two specialized models: PP-FormulaNet-L, tailored for high-accuracy scenarios, and PP-FormulaNet-S, optimized for high-efficiency contexts. Our extensive evaluations reveal that PP-FormulaNet-L attains accuracy levels that surpass those of prominent models such as UniMERNet by a significant 6\%. Conversely, PP-FormulaNet-S operates at speeds that are over 16 times faster. These advancements facilitate seamless integration of PP-FormulaNet into a broad spectrum of document processing environments that involve intricate mathematical formulas. Furthermore, we introduce a \textbf{Formula Mining System}, which is capable of extracting a vast amount of high-quality formula data. This system further enhances the robustness and applicability of our formula recognition model. Code and models are publicly available at PaddleOCR\footnote{\url{https://github.com/PaddlePaddle/PaddleOCR}} and PaddleX\footnote{\url{https://github.com/PaddlePaddle/PaddleX}}.
\end{abstract}

%% file: sec/1_intro.tex
\section{Introduction}
\label{sec:intro}
\begin{figure}
\centering
\subfigure[Definition of Formula Recognition]{
\begin{minipage}[b]{0.45\textwidth} 
\includegraphics[width=1\textwidth]{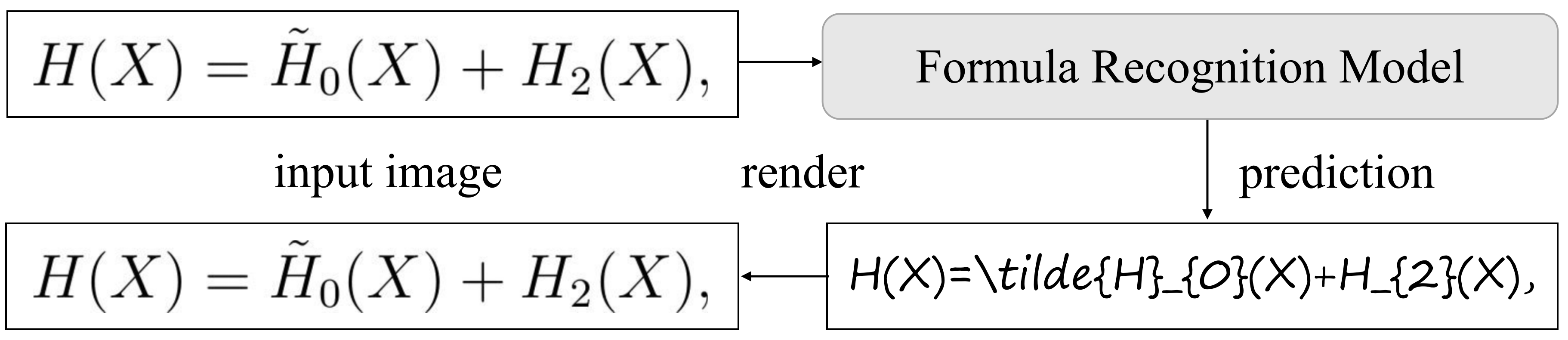}
\end{minipage}
}
\subfigure[Comparison of the Avg-BLEU and FPS of Different Models]{
\begin{minipage}[b]{0.45\textwidth} 
\includegraphics[width=1\textwidth]{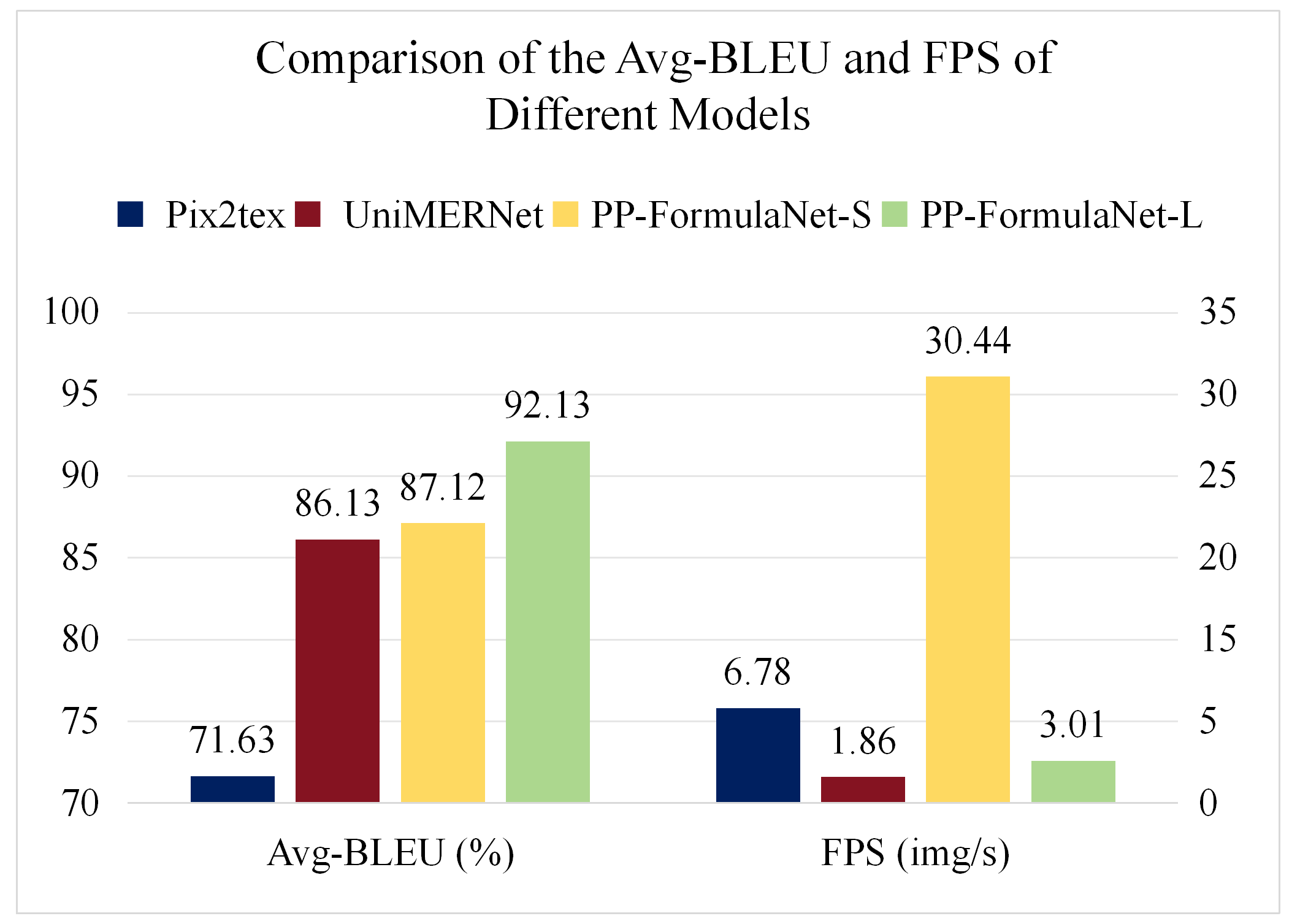}
\end{minipage}
}
\caption{Definition of Formula Recognition, and Comparison of the Avg-BLEU and FPS of Different Models. GPU Inference Time tested on 32G Tesla V100 with batch size of 15.} 
\label{intro_chap3}
\end{figure}
\par Formulas, as the core knowledge carriers in scientific literature, technical documents, and educational materials, embody the abstract logic and mathematical expressions of human civilization. With the development of large language models, multimodal AI, and intelligent scientific computation, the structured parsing of formula semantics has become a critical breakthrough for constructing scientific knowledge graphs and enhancing AI’s mathematical reasoning capabilities. As shown in Figure \ref{intro_chap3}, utilizing formula recognition technology to structurally parse inter-line formulas, inline formulas, and handwritten formulas in contexts such as academic papers and engineering drawings—combined with symbolic semantic understanding and LaTeX generation capabilities—can infuse large model training with precise mathematical prior knowledge. This significantly enhances AI’s problem-solving, theorem derivation, and scientific document generation abilities in the research field. Additionally, formula recognition can be integrated with large models to facilitate the construction of knowledge bases in the scientific domain.

\par In recent years, thanks to the powerful sequence modeling capabilities of transformers, formula recognition algorithms have developed rapidly. However, most algorithms have a limited vocabulary size of only 300 classes and are focused on the singular scenario of handwritten formula recognition, making them difficult to apply to real-world document parsing scenarios. Additionally, these algorithms tend to prioritize improvements in formula accuracy, often at the expense of inference speed and model size. Therefore, enhancing the generalization ability of formula algorithms in complex scenarios, while optimizing inference speed and model size, holds significant research value.

\par  To address the aforementioned issues, UniMERNet \cite{UniMERNet} uses large-scale pre-training datasets from arXiv to enhance model generalization for complex formulas. Strategies like reducing model dimensions, replacing attention modules with large-kernel convolutions, and using multi-line bidirectional decoding \cite{LAST} improve inference speed. However, these methods have limitations. Crawling programs that rely on \texttt{\textbackslash begin\{equation\}} tags often discard formulas if rendering fails, missing complex structures defined with commands like \texttt{\textbackslash def} and \texttt{\textbackslash newcommand}. Moreover, reducing model size doesn’t always enhance inference speed due to the autoregressive nature of models, where multiple iterative steps are required to predict the formula. However, during model lightweighting, there is an inevitable loss of accuracy, making it challenging for the model to accurately predict the sequence termination point. Consequently, although the number of steps for single inference has decreased, the total number of inference steps has actually increased. Balancing both, there is little change in the overall inference speed of the model.
\par In this paper, we adopt the method of crawling arXiv papers to obtain a large dataset of formula training data. Unlike previous approaches, we develop a more comprehensive formula extraction script that handles user-defined commands and ensured compatibility with various formula identifiers. Based on this program, a dataset of 4 million arXiv formulas is constructed. Additionally, we significantly enhance the CNN network’s capability to model formula structures by distilling the weights of the multi-modal large model GOT2.0-OCR \cite{GOT} onto PP-HGNetV2-B4\cite{PPHGNetv2}. Meanwhile, considering the issue that the original pre-trained weights cannot be loaded after model dimensionality reduction, we propose a dimensional interpolation technique to adjust the pre-trained weights. Finally, we introduce a multi-token prediction method that significantly reduces the number of steps in formula sequence inference, thereby improving inference efficiency.
\par In summary, this work developed an accurate and comprehensive formula mining system and proposed PP-FormulaNet based on techniques such as weight interpolation, knowledge distillation, and multi-token prediction, to meet the performance requirements of practical document processing scenarios. The main contributions are as follows:

\begin{itemize}
   \item We  develop an innovative formula mining system that considers various potential formula identifiers and restores complex formulas containing user-defined commands. This significantly enhances the completeness and accuracy of formula dataset construction.
    \item We introduce PP-FormulaNet, a formula recognition model specifically designed for real-world document analysis scenarios. As depicted in Figure \ref{intro_chap3}, \textbf{PP-FormulaNet-L} delivers precision levels that are \textbf{6\%} higher than those of leading models such as UnimerNet, while \textbf{PP-FormulaNet-S} functions at speeds that are \textbf{16} times faster.
     \item We design methods such as weight interpolation and model distillation to effectively utilize the formula representations in pre-trained weights during the model lightweighting process.
      \item We offer a new perspective on accelerating formula inference by employing multi-token prediction techniques, which significantly reduce the number of inference steps for formula sequences.
\end{itemize}

\section{Related Works}

\par Existing formula recognition algorithms can mainly be classified into three categories: traditional machine learning methods, deep learning methods, and large model-driven methods. 
\subsection{Traditional Machine Learning Methods}

\par  Traditional machine learning methods \cite{formula_recognition_rule1, formula_recognition_rule3, formula_recognition_rule4} primarily rely on handcrafted feature extraction and classifiers. These methods typically consist of three main steps: character grouping, character recognition and relationship analysis. \cite{formula_recognition_rule4} was the first to use a set of replacement rules to model the two-dimensional structure of formula images. \cite{formula_recognition_rule3}  improved the accuracy of formula character recognition by using segmental K-means to initialize the Gaussian Mixture Model parameters and introduced features such as the normalized distance of stroke edges to model formula characters. \cite{formula_recognition_rule1} used Hidden Markov Models (HMM) to recognize formula characters and employed stochastic context-free grammar to model the structural relationships between characters.  While these methods achieved some success in early formula recognition tasks, their generalization ability and robustness are limited due to the reliance on handcrafted features, especially when dealing with complex formulas or images with significant noise.
\begin{figure*}
  \centering
  \includegraphics[width=\linewidth]{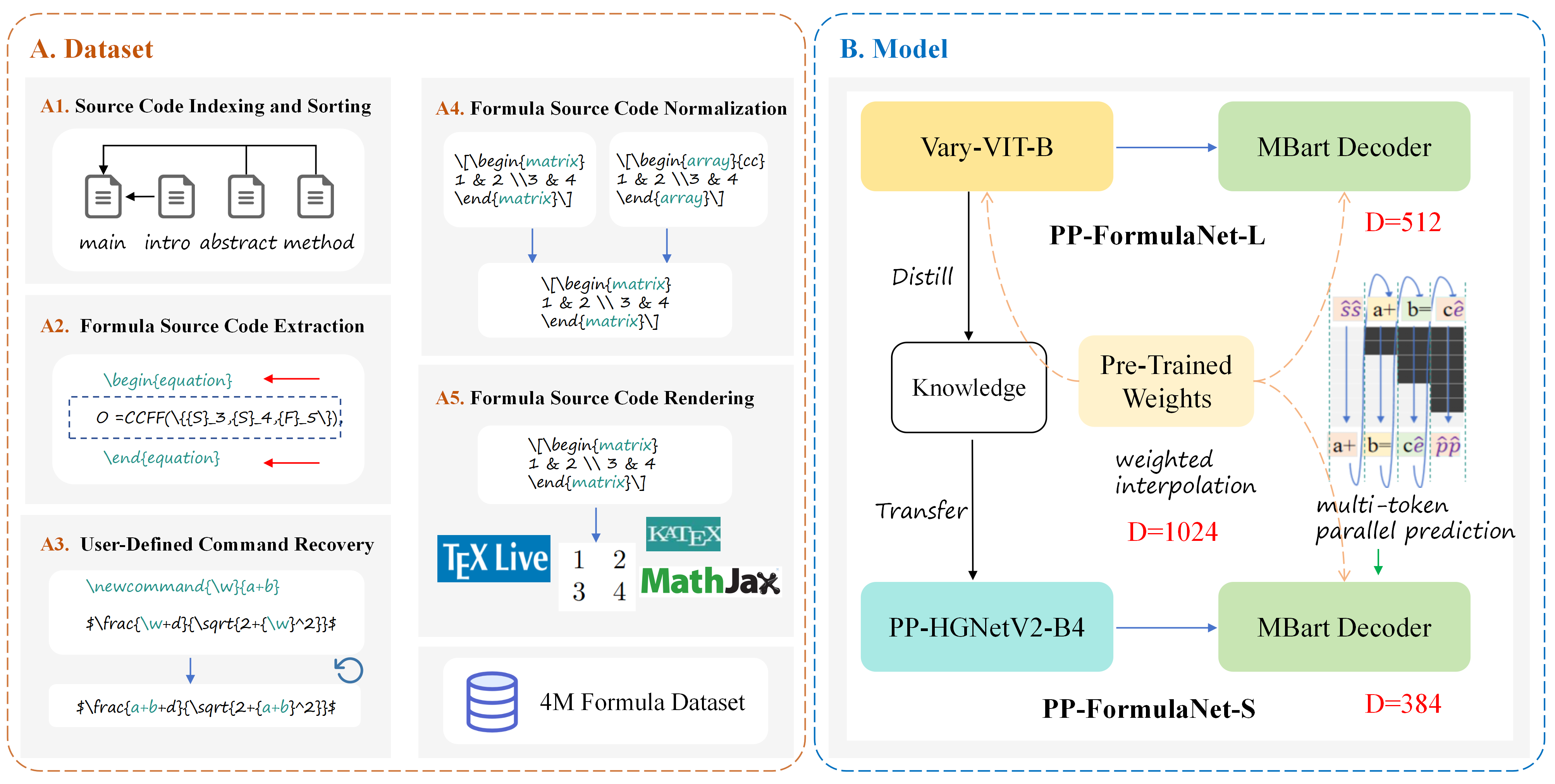}
    \caption{The overall architecture of PP-FormulaNet. Initially, formula-image pairs are extracted from arXiv paper sources through a formula mining system, resulting in a dataset comprising 4 million formulas. PP-FormulaNet consists of two variants:
(1) PP-FormulaNet-L: A vision encoder based on the Vary-VIT-B backbone from GOT-2.0, combined with a 512-dimensional MBart Decoder.
(2) PP-FormulaNet-S: A vision encoder based on the distilled PP-HGNetV2-B4, combined with a 384-dimensional MBart Decoder. To maintain the formula representation capability of pre-trained weights, weight interpolation is applied to adjust the Vary-VIT-B resolution from $1024\times1024$ to $768\times768$, and the decoder dimensions from 1024 to 512/384. Additionally, multi-token prediction is implemented in PP-FormulaNet-S to enhance inference speed by predicting multiple consecutive tokens in a single forward pass.  $\widehat{s}$, $\widehat{e}$ and $\widehat{p}$ represent the start token, end token, and padding token, respectively.}
  \label{framework}
\end{figure*}
\subsection{Deep Learning Methods}
\par In recent years, with the rapid development and widespread application of deep learning technology, the field of formula recognition has achieved breakthrough progress, with significant improvements in both accuracy and recognition efficiency. Current deep learning-based methods \cite{WAP,DWAP,WS_WAP,CAN,BTTR, ABM,LAST,CoMER,SAN} typically employ an encoder-decoder architecture. The encoder encodes the features of the formula image into visual tokens, and subsequently, the decoder applies an attention mechanism to these visual tokens to obtain the decoded formula sequence. The WAP \cite{WAP} uses convolutional networks to encode formula images and recurrent neural networks to decode the encoded features. It introduces a coverage attention mechanism to address the issues of over-parsing and under-parsing during the formula decoding process. Building on this foundation, the DWAP\cite{DWAP} and WS-WAP\cite{WS_WAP} enhance visual features by introducing multi-scale attention and a symbol classifier into the encoder, respectively. The CAN \cite{CAN} addresses the issue of inaccurate attention when decoding complex formulas by introducing a character counting task during formula recognition training. BTTR \cite{BTTR} and ABM \cite{ABM} utilize the Transformer architecture to replace traditional RNN decoders. By incorporating a bidirectional language modeling mechanism, they significantly enhance the model’s capability to handle long LaTeX expression sequences. LAST \cite{LAST} innovatively employs a line-level bidirectional decoder strategy combined with a line position encoding module to achieve parallel bidirectional decoding of multi-line mathematical formulas. CoMER \cite{CoMER} introduces an attention refinement module and divides coverage attention into self-coverage and cross-coverage to address the issue of lacking coverage information in transformer-based formula recognition models. SAN \cite{SAN} defines a systematic set of syntactic rules to convert LaTeX sequences into structured parse tree representations. By reformulating the sequence prediction task as a traversal process of the parse tree, the model significantly reduces the error rate in formula prediction. Existing methods have primarily been optimized for handwritten formula recognition tasks, often underperforming when dealing with printed formulas. To address the challenges of formula recognition in real document parsing scenarios, researchers have introduced innovative methods such as pix2tex \cite{pix2tex}, texify \cite{Texify}, and UniMERNet \cite{UniMERNet}. Notably, UniMERNet \cite{UniMERNet} has been trained on a large-scale dataset consisting of millions of samples, significantly enhancing the model’s recognition performance across various scenarios.

\subsection{Large Model-Driven Methods}
\par Amidst the rapid advancements in large model technologies, researchers are focusing on transferring the pre-training paradigm of large-scale vision-language models to the domain of formula recognition.  Donut \cite{Donut} and Nougat \cite{Nougat} demonstrate significant advantages in understanding documents, and excel in capturing the overall structure and content of entire documents. On the other hand,  Vary \cite{Vary} and GOT-OCR2.0 \cite{GOT} focus on more fine-grained document analysis and perception tasks, allowing for detailed examination and processing of specific document components. At the same time, general-purpose multimodal large models, exemplified by Qwen2.5-VL \cite{Qwen2.5-VL} and InternVL2.5 \cite{InternVL_2_5B}, show exceptional performance in formula recognition. These models are capable of effectively handling the parsing requirements of most complex formulas. However, due to the large scale of model parameters, the inference efficiency of multimodal models is significantly lower than that of traditional deep learning inference methods. This lack of computational efficiency severely limits their practical applicability in formula-intensive document parsing scenarios.
\section{Approach}

\par The overall architecture of PP-FormulaNet is illustrated in Figure \ref{framework}. A formula extraction system is employed to gather a substantial dataset of 4 million formula source code and image pairs from the source code of arXiv papers. The backbone network Vary-VIT-B from GOT2.0-OCR \cite{GOT} is utilized to develop PP-FormulaNet-L, serving as the visual encoder, along with an MBart Decoder with a dimension of 512 as the decoder. For PP-FormulaNet-S, the PP-HGNetv2-B4\cite{PPHGNetv2}, distilled from Vary-VIT-B, is used as the visual encoder, paired with an MBart Decoder with a dimension of 384. Weight interpolation techniques are applied to adjust the Vary-VIT-B weights, initially designed for a resolution of $1024\times1024$, down to $768\times768$. Additionally, the MBart decoder dimensions are interpolated from the original 1024 to 512 and 384, ensuring maximal retention of pre-trained weights for formula representation. Lastly, a multi-token prediction technique is implemented to optimize the speed of PP-FormulaNet-S, allowing it to predict multiple consecutive tokens simultaneously during inference.

\subsection{Formula Mining System}
\par 
Existing formula extraction methods typically rely on parsing specific markup (such as \texttt{\textbackslash begin\{equation\}} and \texttt{\textbackslash end\{equation\}}) within the source code of papers to extract formulas. They then use rendering engines like KaTeX or pdflatex to convert the formula source code into images. However, the source code for formulas in real documents often has complex structures and diverse formats, such as custom macros, nested commands, or dynamically generated content. This complexity limits traditional methods to extracting formulas in simpler forms, making it difficult to comprehensively cover the complex formula expressions found in real-world scenarios.
\par To address the aforementioned issues, we design an innovative formula extraction system. This system takes into account various formula identifiers and user-defined commands that may appear in real documents. It primarily consists of the following five steps:
\par \textbf{1) Source Code Indexing and Sorting}

The order of the source code is crucial for the proper parsing of formulas, particularly because user-defined commands are typically defined at the beginning of the source code. Ensuring the correct order of the source code is essential; otherwise, these custom commands cannot be accurately restored using regular expressions. However, the actual arXiv source code often includes multiple source files with complex inter-file referencing relationships. For instance, a file like main.tex might need to reference abstract.tex and Intro.tex. 
To address this issue, the main tex file is first identified using identifiers such as \texttt{\textbackslash usepackage}, and then the order of the source code is restored based on the main tex file and reference identifiers.

\par \textbf{2) Formula Source Code Extraction}
\par Previous methods typically extract formulas using simple identifiers such as \texttt{\$\$} or \texttt{\textbackslash begin\{equation\}}, resulting in a limited variety of extracted formula types. To extract a richer and more diverse range of formulas, an in-depth analysis of formula identifiers in arXiv papers is conducted. First, figures and tables are filtered out from the source code since arXiv papers often contain numerous formulas coupled with figures and tables (such as inter-figure and inter-table formulas). These formulas are intermingled with the figure and table source code, making them difficult to separate and potentially interfering with the algorithm’s ability to learn formula features. Next, the list of formula identifiers is expanded to include additional identifiers such as \texttt{\textbackslash begin\{align\}}, \texttt{\textbackslash begin\{align*\}}, \texttt{\textbackslash begin\{multline\}}, \texttt{\textbackslash begin\{gather\}}, and \texttt{\textbackslash begin\{eqnarray\}}. Finally, using this expanded list of identifiers, a comprehensive extraction of formula source code is conducted, enabling a more accurate capture of a diverse range of formula structures.

\par \textbf{3) User-Defined Command Recovery}
\par In actual arXiv papers, authors often use commands like \texttt{\textbackslash DeclarePairedDelimiter}, \texttt{\textbackslash newcommand}, and \texttt{\textbackslash DeclareMathOperator} to customize formula source code, such as substituting \texttt{\textbackslash eps} for \texttt{\textbackslash varepsilon}. These custom commands do not conform to standard LaTeX syntax, rendering formulas that include them unable to be directly processed. To address this issue, a series of regular expression rules were designed to parse and restore these custom commands. For definitions such as \texttt{\textbackslash newcommand{x}{\textbackslash mathbf{x}}}, the mapping between custom commands and actual LaTeX code is first stored in a hash table. Regular expressions are then used to match all instances containing the command and identify their starting positions. A stack algorithm is employed to determine their ending positions. Finally, based on the starting and ending positions and the regular expression rules, custom commands are replaced with standard LaTeX code.

\par \textbf{4) Formula Source Code Normalization
}
When processing mathematical formula source code scraped from arXiv, a significant amount of ambiguity is encountered, posing notable challenges for subsequent machine learning tasks. A typical example is the different LaTeX representations for visually identical matrix structures. For instance, a matrix can be defined using either the \texttt{\textbackslash begin\{array\}} or the \texttt{\textbackslash begin\{matrix\}}. Although they produce the same visual result when rendered, these different representations introduce unnecessary noise at the data level, potentially affecting the model’s training performance. To address this issue, a systematic solution is implemented by employing KaTeX to normalize the source code. KaTeX, a fast and reliable library for rendering LaTeX in web environments, provides a consistent way to parse and render mathematical expressions. By using KaTeX, different but equivalent LaTeX syntax is converted into a standardized form, thus reducing variability and ambiguity in the dataset.

\par \textbf{5) Formula Source Code Rendering}
\par After cleaning the formula source code, the local rendering engine pdflatex is used to render the formulas, converting the formula source code into formula images and thus constructing source code-image pairs. This process involves three steps: first, embedding the formula into a LaTeX template to generate a .tex file. Next, pdflatex renders the .tex file into a PDF containing the formula. Finally, the fitz tool is utilized to convert the PDF into page images, and OpenCV algorithms are employed to crop out the formula regions.

\subsection{Weighted Interpolation}
\par Pre-trained models are essential for enhancing the generalization capabilities of formula recognition models. Experiments demonstrate that models loaded with pre-trained weights achieve significantly higher recognition accuracy than those trained from scratch. However, compressing the formula recognition model necessitates trimming model dimensions (e.g., reducing from 1024 dimensions to 512 or 384), which prevents directly loading the original pre-trained weights. Notably, existing pre-trained models for formula recognition, such as Donut \cite{Donut} and GOT2.0-OCR \cite{GOT}, typically require training on datasets with millions of samples using dozens of GPUs for hundreds of hours, making it unfeasible to re-pretrain the trimmed models in practical scenarios. To overcome this challenge, an innovative weight interpolation method is proposed, which adaptively adjusts the original pre-trained weights, enabling the trimmed model to effectively load them. This approach maintains the model’s lightweight nature while fully leveraging the benefits of pre-trained models.

\begin{table*}
    \centering
       \caption{The Recognition Performance of Different Methods on the UniMERNet-1M and  arXiv-4M Test Sets.  } \label{compare_exp}
         
    \begin{tabular}{c|cc|ccc|c|c|c} 
        \hline
        \multirow{3}{*}{Model}  & \multicolumn{2}{c|}{UniMERNet-1M} &  \multicolumn{3}{c|}{arXiv-4M} &\multirow{3}{*}{\shortstack{Avg-\\BLEU$\uparrow$}}   & \multirow{3}{*}{ \shortstack{GPU inference\\ time (ms)\\(batch=1)}}&  \multirow{3}{*}{ \shortstack{GPU inference\\ time (ms)\\(batch=15)}}\\
        \cline{2-6}
        &\shortstack{SPE-\\BLEU$\uparrow$} & \shortstack{CPE-\\BLEU$\uparrow$}&
         \shortstack{Easy-\\BLEU$\uparrow$} &\shortstack{Middle-\\BLEU$\uparrow$} & \shortstack{Hard-\\BLEU$\uparrow$} & &
         
        \\
        \hline
    Pix2tex\cite{pix2tex}& 0.8780& 0.5670& 0.8151&0.7546 &0.5746&0.7163 & \cellcolor{cyan!15}\underline{1244.61} & \cellcolor{cyan!15}\underline{147.38} \\
    UniMERNet \cite{UniMERNet}&\cellcolor{red!15}\textbf{0.9187}&\cellcolor{red!15}\textbf{0.9251}&0.8659&0.8229&0.7741&0.8613&2266.96&536.75 \\
    \hline
    PP-FormulaNet-S &0.8694&0.8071& \cellcolor{cyan!15}\underline{0.9295}& \cellcolor{cyan!15}\underline{0.9113}& \cellcolor{cyan!15}\underline{0.8392}& \cellcolor{cyan!15}\underline{0.8712}&\cellcolor{red!15}\textbf{202.25}&\cellcolor{red!15}\textbf{32.85}\\
    PP-FormulaNet-L & \cellcolor{cyan!15}\underline{0.9055}& \cellcolor{cyan!15}\underline{0.9207}&\cellcolor{red!15}\textbf{0.9392}&\cellcolor{red!15}\textbf{0.9273}&\cellcolor{red!15}\textbf{0.9142}& \cellcolor{red!15}{\textbf{0.9213}}&1976.52&332.12\\
    \hline
    \end{tabular}
\end{table*}

\par To address the challenge of adapting pre-trained models to reduced dimensions, an attention module typically composed of linear layers and normalization layers can be optimized. The key assumption here is that adjacent dimensions of attention weights in the decoder encode similar semantic information. This assumption permits the mixing of dimensional features through interpolation, allowing the reduced-dimension model to still effectively load pre-trained weights. For the linear layers and normalization layers, the calculation method for the weights after dimension reduction is as follows:
\begin{equation}
\left\{\begin{array}{c}
F_{\text {Linear }}^D=\varphi\left(F_{\text {Linear }}\right) \\
F_{\text {norm }}^D=\tau_2\left(\varphi\left(\tau_1\left(F_{\text {norm }}\right)\right)\right)
\end{array}\right.
\end{equation}
\par  Here,  $ F_{\text {Linear }}^D$ and  $F_{\text {norm }}^D$ respectively represent the original pre-trained weights of the linear layer and the normalization layer. $F_{\text {Linear }}$ and $F_{\text {norm }}$ denotes the interpolated pre-trained weights of the linear layer and the normalization layer.
$\varphi$ denotes the nearest neighbor interpolation, $\tau_1$ represents the unsqueeze operation used to reshape the dimensions of the normalization layer from $\mathbb{R}^C$ to $\mathbb{R}^{C \times 1}$, and $\tau_2$ represents the squeeze operation used to reshape the dimensions of the normalization layer from $\mathbb{R}^{C \times 1}$ to $\mathbb{R}^{C}$.

\subsection{Knowledge Distillation for PP-FormulaNet-S Backbone}
\par For PP-FormulaNet-S, its backbone network faces challenges in adapting to formula recognition tasks due to limited parameter capacity and absence of domain-specific knowledge in mathematical notation. To address this, we employ knowledge distillation to transfer capabilities from Vary-VIT-B. The teacher network parameters remain frozen while training PP-HGNetV2-B4, leveraging feature-level supervision through a fully-connected layer. 

Let $\mathbf{F}_{\text{Vary-VIT-B}} \in \mathbb{R}^{B \times D}$ and $\mathbf{F}_{\text{PP-HGNetV2-B4}} \in \mathbb{R}^{B \times P}$ denote the feature tensors from teacher and student networks respectively, where $D$ and $P$ represent their corresponding feature dimensions, $B$ denotes the Batch size. To bridge the dimensional discrepancy ($D \neq P$), we introduce a learnable linear projection $\varphi: \mathbb{R}^{P} \to \mathbb{R}^{D}$. The distillation loss is formulated as:

\begin{equation} 
\mathcal{L}_{\text{Distill}} = \frac{1}{B} \sum_{i=1}^{B} \left\lVert \mathbf{F}_{\text{Vary-VIT-B}}^{(i)} - \varphi(\mathbf{F}_{\text{PP-HGNetV2-B4}}^{(i)}) \right\rVert_{2}^{2}
\end{equation}

The distillation framework was trained on a diverse corpus containing 500000 document samples spanning five domains: 
\begin{itemize}
    \item Mathematical formulas (including equation derivations and symbolic notations)
    \item Financial documents (reports and balance sheets)
    \item Scientific literature (arxiv papers in STEM fields)
    \item Academic dissertations (with complex layout structures)
    \item Tabular data (statistical reports and spreadsheets)
\end{itemize}

Training was conducted at $768\times$768 resolution for 50 epochs using AdamW optimizer ($\beta_1=0.9$, $\beta_2=0.999$). The distilled PP-HGNetV2-B4 achieves effective feature extraction capabilities with only 15.6 M parameters.

This compact model learns transferable representations for OCR tasks, elevating the formula recognition BLEU score from 80.87 to 84.32 (+3.43pp) compared to ImageNet-pretrained baselines.

\subsection{Multi-Token Parallel Prediction
}
\par Current mainstream formula recognition algorithms typically employ an autoregressive generation architecture, which involves predicting the formula sequence token by token. This approach can significantly reduce inference efficiency, particularly when dealing with complex formulas that exceed 1024 tokens in length. While existing studies, such as the LAST method \cite{LAST}, introduced multi-line parallel inference strategies to enhance efficiency, these strategies are generally confined to scenarios involving multi-line formulas and still encounter challenges with complex structures like matrices and braces. To address these limitations, an innovative multi-token parallel prediction method is proposed, designed to improve the inference efficiency of complex formulas.
\par 
As shown in the Fig. \ref{framework}, the parallel causal mask is a key component of the multi-token prediction technique, and its mathematical expression is as follows:
\begin{align}
\mathbf{M}_{ij} =
\begin{cases}
0, & \text{if } \left\lfloor \frac{i}{step} \right\rfloor \geq \left\lfloor \frac{j}{step} \right\rfloor\\
\\
-\infty, & \text{if } \left\lfloor \frac{i}{step} \right\rfloor < \left\lfloor \frac{j}{step} \right\rfloor\
\end{cases} 
\end{align}
\par Here, 
 $\mathbf{M}_{ij}$ represents the element in the i-th row and j-th column of the parallel causal mask, and 
$step$ indicates the number of tokens predicted at each step. $\left\lfloor \quad \right\rfloor $ represents the floor operation.
\par By introducing the parallel causal mask mechanism, the decoder is equipped to perform multi-step predictions during training: it begins by predicting characters from position step + 1 to 2step, using the initial step characters as a basis. Next, it predicts characters from position 2step + 1 to 3step, relying on the first 2step characters, and continues this process until the maximum length of the current data batch is reached. During the inference phase, the initial characters are replicated step times as input. The model then predicts the next sequence, also of length step, based on this initial sequence, thus achieving multi-token parallel prediction.
\section{Experiment}
\subsection{Datasets and Evaluation Metrics}
\par In this paper, the model is trained on the UniMERNet-1M dataset \cite{UniMERNet} and a self-constructed arXiv-4M dataset, totaling 5 million entries. The proposed method is evaluated through comparative experiments across multiple test sets. Specifically, the test sets include the simple and complex formula test sets from UniMERNet \cite{UniMERNet}, as well as the simple, medium, and complex formula test sets from arXiv-4M. In line with most formula recognition methods, the BLEU-Score \cite{BLEU_score} on the validation set is used as the evaluation metric.
\subsection{Implementation Details}
\par During the training phase, a data augmentation strategy consistent with UniMERNet \cite{UniMERNet} is employed, which specifically includes operations such as dilation, erosion, weather noise, random affine transformations, Gaussian noise, and random brightness and contrast adjustments. For optimization, the AdamW optimizer is utilized with parameters set as beta1=0.9, beta2=0.999, and a weight decay of 0.05. The batch size per GPU is set to 6, and the total number of training epochs is 10. To ensure that the network can progressively learn the features of formulas, we employ a WarmUp strategy with a learning rate of 1e-5 for the first 5,000 steps to preheat the training process, followed by a cosine annealing strategy with a learning rate of 1e-4 to dynamically adjust the learning rate.

\subsection{Comparison with State-of-the-Arts}
\par Comparative experiments were conducted with state-of-the-art (SOTA) methods on the test sets of the arXiv-4M dataset and UniMERNet-1M, as detailed in Table \ref{compare_exp}. In terms of accuracy, PP-FormulaNet-S and PP-FormulaNet-L achieve an average BLEU score (Avg-BLEU) improvement of 0.99\% and 6\%, respectively, over the current open-source SOTA method, UniMERNet\footnote{We are comparing the original version of UniMERNet. We will provide a more detailed comparison with the new version in the future.}, across five test datasets, demonstrating significant performance gains. Regarding inference speed, under batch=15, PP-FormulaNet-S achieves a 16x faster GPU inference speed compared to UniMERNet, while PP-FormulaNet-L reduces the inference time by 61\%. These results indicate that PP-FormulaNet not only surpasses existing methods in accuracy but also achieves substantial improvements in inference efficiency. This is particularly evident in batch processing scenarios, where its performance advantages are even more pronounced.

\subsection{Ablation Study}
\par Ablation experiments were conducted on the UniMERNet-1M dataset to verify the impact of knowledge distillation, weight interpolation, and multi-token parallel prediction on formula recognition performance.
\begin{table}
    \centering
    \caption{The Impact of Weight Interpolation on Recognition Performance.} \label{weight_interpolation}
    
    \begin{tabular}{c|c|cc}
       \hline
       \multirow{1}{*}{Backbone}& Dim    & \makecell{Weight \\Interpolation} &\makecell{ CPE-\\BLEU$\uparrow$}    \\ 
          \hline
           \multirow{3}{*}{Donut-Swin\cite{UniMERNet}}&1024 & &0.8968\\
           \cline{2-4}
          & \multirow{2}{*}{512}& \cellcolor{gray!20}&\cellcolor{gray!20}0.7970\\
          &&\cellcolor{red!15}\checkmark&\cellcolor{red!15}\textbf{0.8445}\\
         
        \hline
        Vary-VIT-B \cite{GOT} &512&\checkmark &0.9148\\
           \hline
    \end{tabular}
\end{table}

\par \textbf{Impact of weight interpolation} 
To enable the dimension-pruned model to continue utilizing pre-trained weights, a weight interpolation technique is proposed. Experimental results in Table \ref{weight_interpolation} demonstrate that this method effectively preserves the performance advantages of the pre-trained model. Specifically, when the recognition network is UniMERNet, using pre-trained weights results in a 4.75\% performance improvement on the complex formula dataset, fully validating the effectiveness of the weight interpolation technique. However, despite the substantial performance improvement achieved by the weight interpolation technique, a 5.23\% performance gap remains compared to the original UniMERNet model with a dimension of 1024. To further optimize model performance, replacement experiments on the backbone network were conducted. The results indicate that when Vary-VIT-B is used as the backbone network, the model’s recognition performance reaches 0.9148, representing a 1.8\% improvement compared to the original UniMERNet with a dimension of 1024. Based on this outcome, Vary-VIT-B is ultimately selected as the backbone network for PP-FormulaNet-L.

\par \textbf{Impact of knowledge distillation}

\begin{table}
    \centering
    \caption{The Impact of Knowledge Distillation on Recognition Performance Across Different Backbone Networks.} \label{knowledge_distill}
    
    \begin{tabular}{c|c|c@{\hskip 0.2cm}c|c}
       \hline
       \multirow{1}{*}{Backbone}    & \makecell{Knowledge \\Distillation}& \makecell{ SPE-\\BLEU$\uparrow$}  &\makecell{ CPE-\\BLEU$\uparrow$} & Params   \\ 
          \hline
          \multirow{2}{*}{ResNet50\cite{ResNet50}}& &0.7871&0.5737&\multirow{2}{*}{23.6 M}\\
          &\checkmark&0.8107&0.6216& \\
        \hline
   \multirow{2}{*}{\makecell{PP-HGNetV2\\-B1\cite{PPHGNetv2}}}& &0.5963&0.2703 & \multirow{2}{*}{2.2M}\\
          &\checkmark&0.7506&0.4851\\
    \hline
       \multirow{2}{*}{\makecell{PP-HGNetV2\\-B4\cite{PPHGNetv2}}}&\cellcolor{gray!20} & \cellcolor{gray!20}0.8087&\cellcolor{gray!20}0.6357& \multirow{2}{*}{15.6M}\\
          &\cellcolor{red!15}\checkmark&\cellcolor{red!15}\textbf{0.8432}&\cellcolor{red!15}\textbf{0.6874}\\
           \hline
    \end{tabular}
\end{table}
Despite the Vary-VIT-B backbone network achieving outstanding recognition performance, its large number of parameters and the inherent $O(n^2)$ computational complexity of Transformers lead to longer inference times. To address this issue, knowledge distillation experiments were conducted on Vary-VIT-B using various CNN-based backbone networks. The experimental results in Table \ref{knowledge_distill} demonstrate that knowledge distillation significantly enhances model performance across different backbone networks. Specifically, when the backbone network is ResNet50, knowledge distillation results in BLEU-Score improvements of 2.36\% and 4.79\% on simple and complex formula datasets, respectively. When the backbone network is PP-HGNetV2-B1, the effect of knowledge distillation is even more remarkable, achieving BLEU-Score improvements of 15.43\% and 21.48\% on simple and complex formula datasets, respectively. Similarly, when the backbone network is PP-HGNetV2-B4, knowledge distillation also performs excellently, with BLEU-Score improvements of 3.45\% and 5.17\% on simple and complex formula datasets, respectively. These results fully demonstrate the effectiveness and versatility of knowledge distillation across different backbone networks. Additionally, given that PP-HGNetV2-B4 achieved the optimal BLEU scores in the recognition task, it is selected as the backbone network for PP-FormulaNet-S.
\par \textbf{Impact of multi-token parallel prediction} To accelerate the inference of autoregressive models, a multi-token parallel prediction technique is designed, enabling the decoder to predict multiple tokens in a single step. As shown in Table \ref{multi-token_parallel}, when the parallel steps are set to 2, 3, 4, and 5, the model’s inference speed increases by 2.05 times, 2.86 times, 3.77 times, and 4.63 times, respectively. However, as the number of parallel steps increases, the model’s accuracy declines more rapidly: when the parallel steps increase from 1 to 2, accuracy decreases by 1.26\%, when the parallel steps increase from 3 to 4, accuracy decreases by 2.86\%, and when it increases from 4 to 5, the decline in accuracy reaches 3.55\%. Considering the trade-off between accuracy and speed, a parallel step count of 3 is ultimately chosen for the PP-FormulaNet-S model.
\begin{table}
    \centering
    \caption{The Impact of Multi-token Parallel Prediction on Recognition Performance.} \label{multi-token_parallel}
    
    \begin{tabular}{c|c|cc}
       \hline
       \multirow{1}{*}{Backbone}&   \makecell{Parralel \\Step}  &  \makecell{ CPE-\\BLEU$\uparrow$} & \makecell{GPU inference\\ time (ms)(CPE)\\(batch=1)}   \\ 
          \hline
           \multirow{5}{*}{ Vary-VIT-B \cite{GOT}}&\cellcolor{gray!20} 1 &\cellcolor{gray!20}0.9092 & \cellcolor{gray!20}2779.76\\
            & 2&0.8876 & 1353.93\\
   & \cellcolor{red!15}3&\cellcolor{red!15}0.8689 & \cellcolor{red!15}969.30\\
          & 4&0.8403&735.58 \\
          &5&0.8048&600.08\\
         
        \hline
    \end{tabular}
\end{table}
\section{Conclusion and Discussion}
\par 
In this paper, we propose an advanced formula recognition model, PP-FormulaNet, aimed at addressing the inefficiencies and inaccuracies of existing methods in formula recognition within real-world document parsing scenarios. Firstly, a more comprehensive formula mining system is designed, significantly enhancing the coverage and diversity of formula data by expanding the list of formula identifiers and supporting the recovery of user-defined commands. Secondly, formula interpolation and model distillation techniques are innovatively combined to efficiently transfer formula representations from pre-trained weights to PP-FormulaNet, thereby enhancing the model’s generalization ability. Finally, a multi-token prediction technique is introduced to effectively accelerate the inference process of autoregressive models. Experimental results on the UniMERNet-1M and the self-constructed arXiv-4M datasets demonstrate that PP-FormulaNet achieves significant improvements in both recognition accuracy and inference efficiency. However, while the proposed multi-token prediction technique improves inference speed, it also results in some accuracy loss. Reducing this accuracy loss in multi-token prediction is a key focus for future research.

%% file: sec/X_suppl.tex
\onecolumn 
\clearpage
\setcounter{page}{1}
\centering
        \Large
        \textbf{Appendix of ``\thetitle''}
\renewcommand{\thefootnote}{\fnsymbol{footnote}}

\renewcommand{\thetable}{\Alph{table}}
\renewcommand{\theequation}{\Alph{equation}}
\renewcommand{\thefigure}{\Alph{figure}}

\setcounter{table}{0}
\setcounter{section}{0}
\setcounter{figure}{0}
\setcounter{equation}{0}

\definecolor{darkgreen}{HTML}{539165}
\newcommand{\increase}[1]{{
  \fontsize{9.5pt}{0.5em}\selectfont({\color{darkgreen}{$\uparrow$~\textbf{#1}}})
}}

\section{The Visualization Results of Different Formula Recognition Methods on the UniMERNet-1M and arXiv-4M Test Sets}
\subsection{Simple Formula}

\begin{figure}[h]
\centering
\subfigure[ Pix2tex \cite{pix2tex} (25 M)]{
\begin{minipage}[b]{1\textwidth} 
{\includegraphics[height=0.08\textheight,width=1\textwidth]{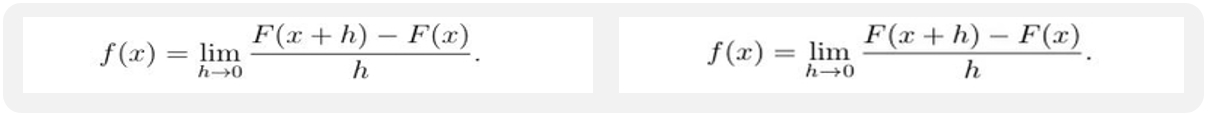}}
\end{minipage}
}
\subfigure[UniMERNet \cite{UniMERNet} (392 M)]{
\begin{minipage}[b]{1\textwidth} 
{\includegraphics[height=0.08\textheight,width=1\textwidth]{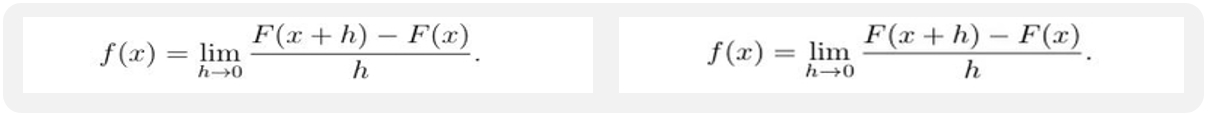}}
\end{minipage}
}
\subfigure[GPT-4o {(Unknown)}]{
\begin{minipage}[b]{1\textwidth} 
{\includegraphics[height=0.08\textheight,width=1\textwidth]{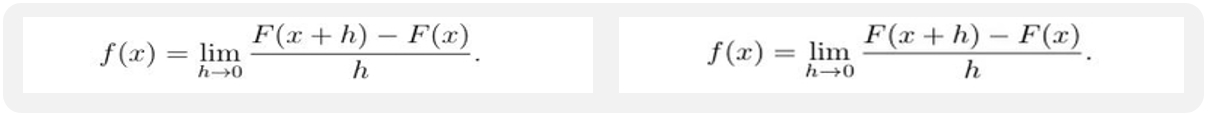}}
\end{minipage}
}
\subfigure[Qwen2.5-VL\cite{Qwen2.5-VL}  {(72B)}]{
\begin{minipage}[b]{1\textwidth} 
{\includegraphics[height=0.08\textheight,width=1\textwidth]{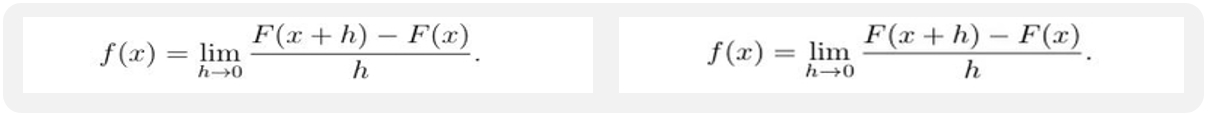}}
\end{minipage}
}
\subfigure[PP-FormulaNet-S (58 M)]{
\begin{minipage}[b]{1\textwidth} 
{\includegraphics[height=0.08\textheight,width=1\textwidth]{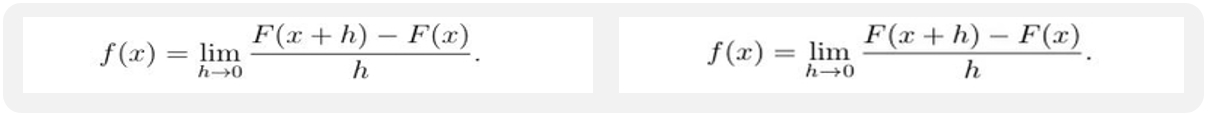}}
\end{minipage}
}
\subfigure[PP-FormulaNet-L (179 M)]{
\begin{minipage}[b]{1\textwidth} 
{\includegraphics[height=0.08\textheight,width=1\textwidth]{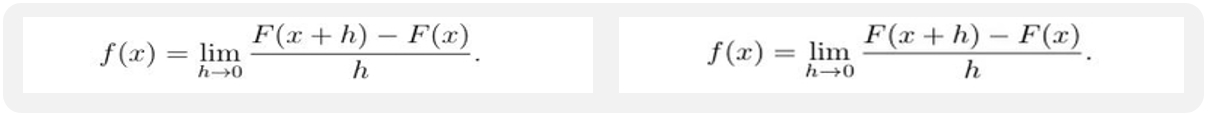}}
\end{minipage}
}
\caption{Visualization of recognition results from different methods on simple formulas. The left image is the test image, and the right image is the prediction result. 
The blank on the right indicates that there is a syntax error in the generated formula, preventing it from rendering.} 
\label{simple_formula}
\end{figure}

\onecolumn 
\clearpage
\setcounter{page}{2}
\subsection{Middle Formula}
\begin{figure}[h]
\centering
\subfigure[Pix2tex \cite{pix2tex} (25 M)]{
\begin{minipage}[b]{1\textwidth} 
{\includegraphics[width=1\textwidth]{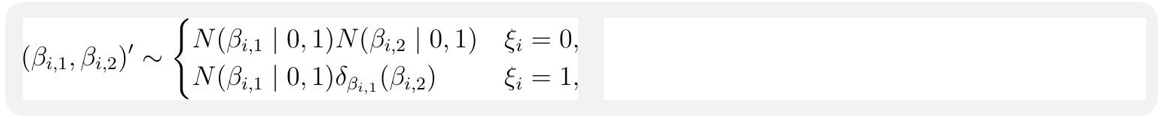}}
\end{minipage}
}
\subfigure[UniMERNet \cite{UniMERNet} (392 M)]{
\begin{minipage}[b]{1\textwidth} 
{\includegraphics[width=1\textwidth]{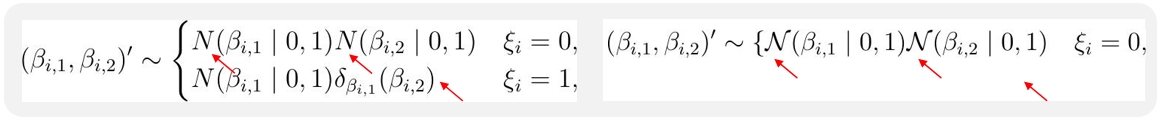}}
\end{minipage}
}
\subfigure[GPT-4o (Unknown)]{
\begin{minipage}[b]{1\textwidth} 
{\includegraphics[width=1\textwidth]{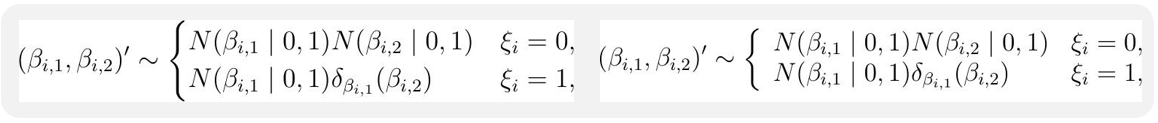}}
\end{minipage}
}
\subfigure[Qwen2.5-VL-72B\cite{Qwen2.5-VL} (72B)]{
\begin{minipage}[b]{1\textwidth} 
{\includegraphics[width=1\textwidth]{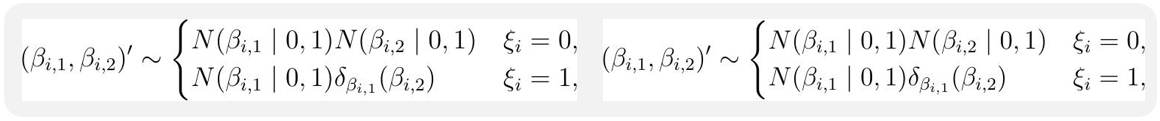}}
\end{minipage}
}
\subfigure[PP-FormulaNet-S (58 M)]{
\begin{minipage}[b]{1\textwidth} 
{\includegraphics[width=1\textwidth]{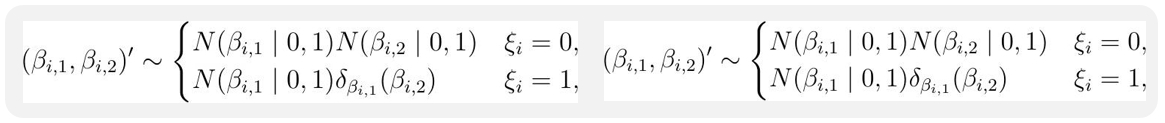}}
\end{minipage}
}
\subfigure[PP-FormulaNet-L (179 M)]{
\begin{minipage}[b]{1\textwidth} 
{\includegraphics[width=1\textwidth]{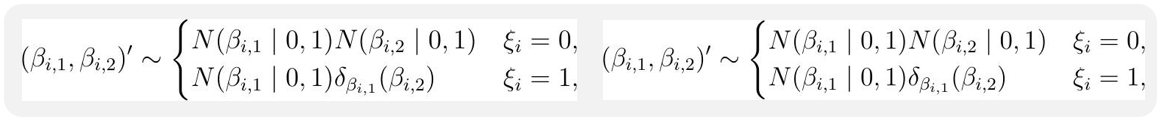}}
\end{minipage}
}
\caption{Visualization of recognition results from different methods on middle formulas. The left image is the test image, and the right image is the prediction result. 
The blank on the right indicates that there is a syntax error in the generated formula, preventing it from rendering.} 
\label{middle_formula}
\end{figure}

\onecolumn 
\clearpage
\setcounter{page}{3}
\subsection{Hard Formula}
\begin{figure}[H]
\subfigure[Pix2tex \cite{pix2tex} (25 M)]{
\begin{minipage}[b]{1\textwidth}
{\includegraphics[width=1\textwidth]{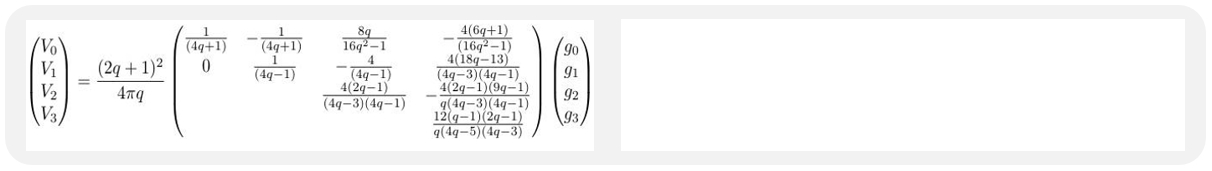}}
\end{minipage}
}
\subfigure[UniMERNet \cite{UniMERNet} (392 M)]{
\begin{minipage}[b]{1\textwidth}
{\includegraphics[width=1\textwidth]{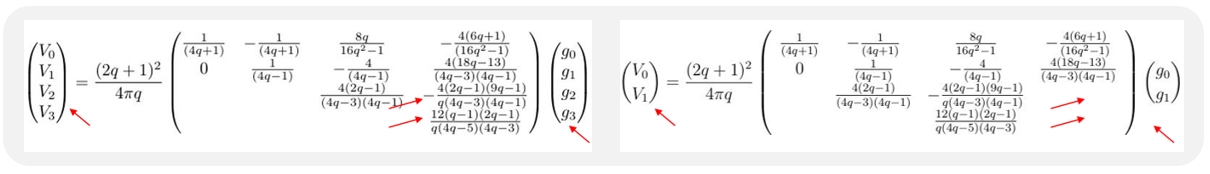}}
\end{minipage}
}
\subfigure[GPT-4o (Unknown)]{
\begin{minipage}[b]{1\textwidth}
{\includegraphics[width=1\textwidth]{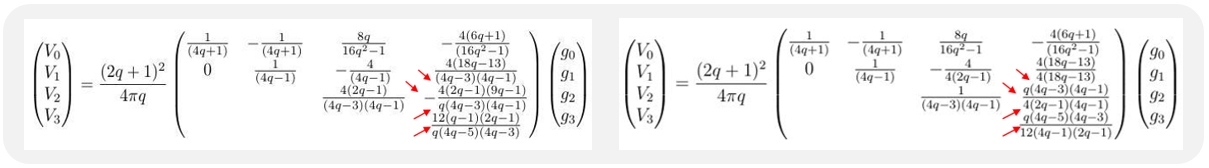}}
\end{minipage}
}
\subfigure[Qwen2.5-VL\cite{Qwen2.5-VL} (72B)]{
\begin{minipage}[b]{1\textwidth}
{\includegraphics[width=1\textwidth]{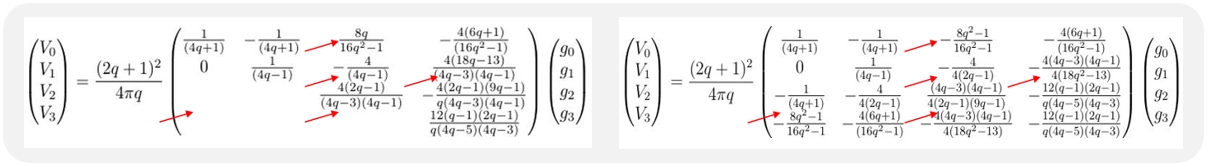}}
\end{minipage}
}
\subfigure[PP-FormulaNet-S (58 M)]{
{\begin{minipage}[b]{1\textwidth}
\includegraphics[width=1\textwidth]{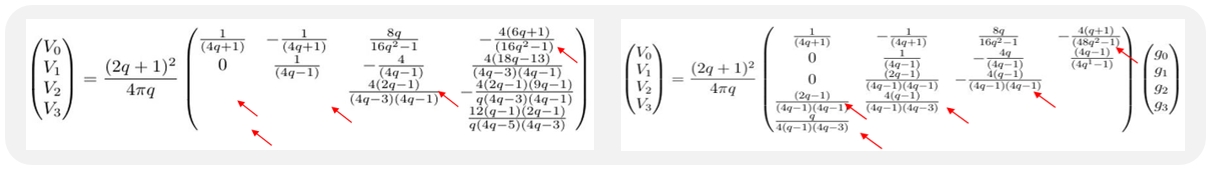}
\end{minipage}}
}
\subfigure[PP-FormulaNet-L (179 M)]{
{\begin{minipage}[b]{1\textwidth}
\includegraphics[width=1\textwidth]{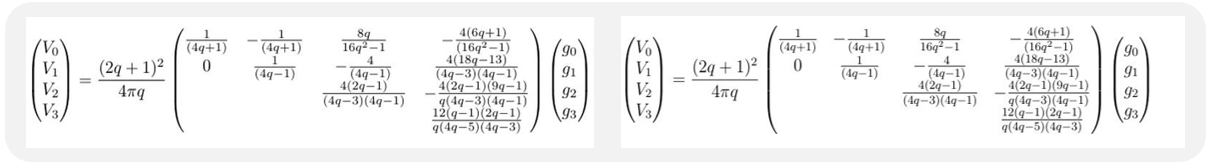}
\end{minipage}}
}
\caption{Visualization of recognition results from different methods on hard formulas. The left image is the test image, and the right image is the prediction result. 
The blank on the right indicates that there is a syntax error in the generated formula, preventing it from rendering.} 
\label{hard_formula}
\end{figure}
\onecolumn 
\clearpage
\setcounter{page}{4}
\subsection{Handwritten Formula}
\begin{figure}[H]
\centering
\subfigure[Pix2tex \cite{pix2tex} (25 M)]{
\begin{minipage}[b]{1\textwidth} 
{\includegraphics[height=0.1 \textheight,width=1\textwidth]{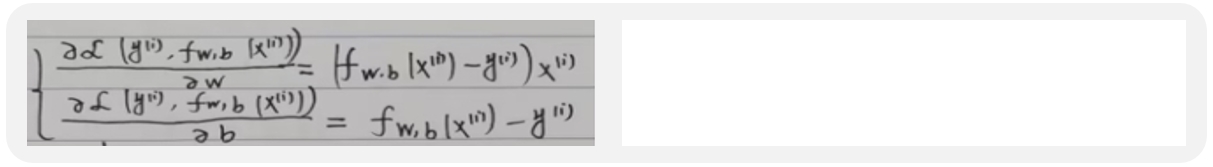}}
\end{minipage}
}
\subfigure[UniMERNet \cite{UniMERNet} (392 M)]{
\begin{minipage}[b]{1\textwidth} 
{\includegraphics[height=0.1 \textheight,width=1\textwidth]{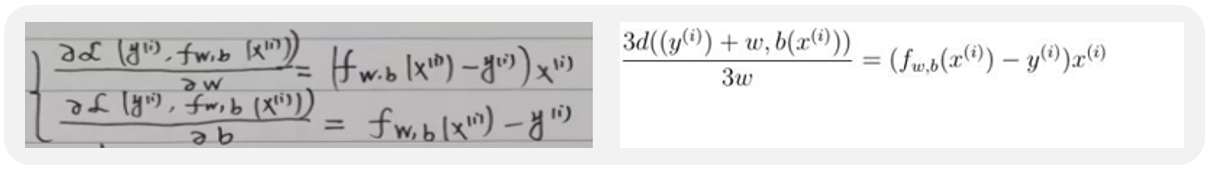}}
\end{minipage}
}
\subfigure[GPT-4o (Unknown)]{
\begin{minipage}[b]{1\textwidth} 
{\includegraphics[height=0.1 \textheight,width=1\textwidth]{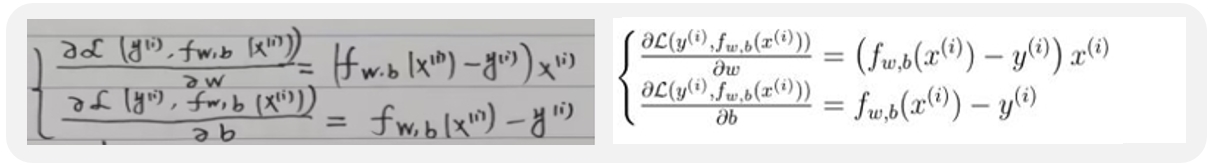}}
\end{minipage}
}
\subfigure[Qwen2.5-VL\cite{Qwen2.5-VL} (72B)]{
\begin{minipage}[b]{1\textwidth} 
{\includegraphics[height=0.1 \textheight,width=1\textwidth]{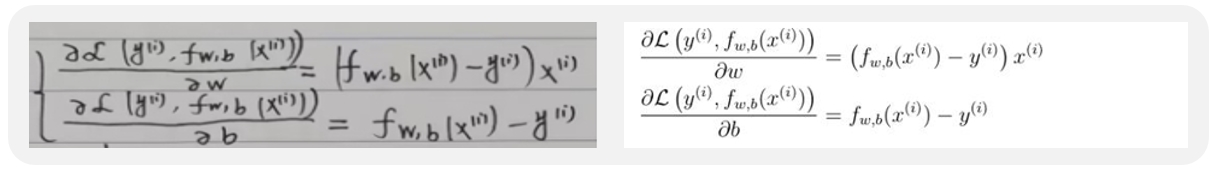}}
\end{minipage}
}
\subfigure[PP-FormulaNet-S (58 M)]{
\begin{minipage}[b]{1\textwidth} 
{\includegraphics[height=0.1 \textheight,width=1\textwidth]{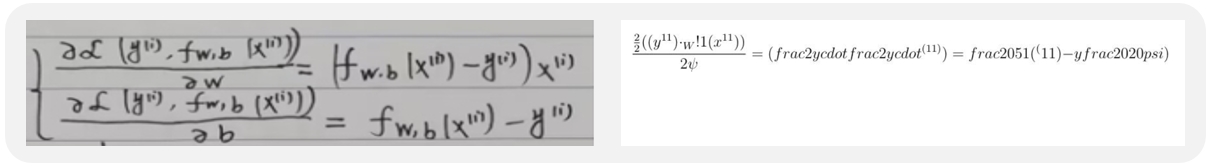}}
\end{minipage}
}
\subfigure[PP-FormulaNet-L (179 M)]{
\begin{minipage}[b]{1\textwidth} 
{\includegraphics[height=0.1 \textheight,width=1\textwidth]{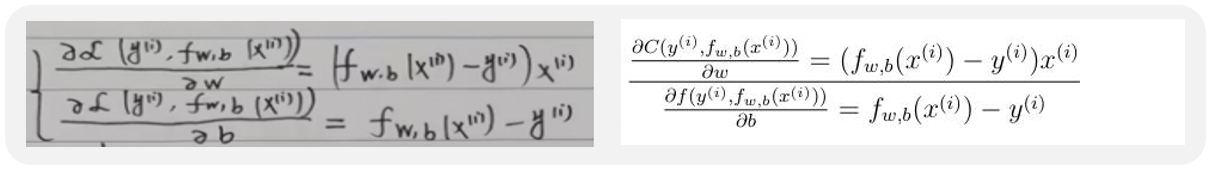}}
\end{minipage}
}
\caption{Visualization of recognition results from different methods on handwritten formulas. The left image is the test image, and the right image is the prediction result. 
The blank on the right indicates that there is a syntax error in the generated formula, preventing it from rendering.} 
\label{handwritten_formula}
\end{figure}

\clearpage
\setcounter{page}{1}